\documentclass{article}
\usepackage{spconf,amsmath,amssymb,graphicx,hyperref, booktabs}
\ninept

\title{Cross-Scale Transfer Learning for Depression Severity Prediction: \\
From PHQ-8 to HAMD-17 Across Languages and Clinical Paradigms}

\name{Wenjie Feng$^{\star}$ \qquad
      Sahba Zojaji$^{\star \dagger \S}$ \qquad
      Satoshi Nakamura$^{\star}$}
\address{$^{\star}$ School of Artificial Intelligence, \\
         $^{\dagger}$ School of Humanities and Social Science, \\
         The Chinese University of Hong Kong, Shenzhen, Shenzhen, China \\
         $^{\S}$ Shenzhen Loop Area Institute, Shenzhen, China}
\begin{document}
%
\maketitle
\vspace{-3mm}
\begin{abstract}    
\vspace{-1mm}
This work addresses continuous depression-severity score prediction from
clinical interview transcripts under data scarcity. We propose a sequential
low-rank adaptation (LoRA) protocol for cross-scale transfer: a Qwen3 backbone
with a bounded regression head is first fine-tuned on the English DAIC-WOZ
dataset (189 avatar-mediated sessions, PHQ-8), and the adapter then initializes fine-tuning on the Chinese PDCH dataset (100 real
clinical consultations, HAMD-17), where a reinitialised, scale-specific head
predicts the clinician-assigned score. All configurations use patient-level stratified 5-fold, 2-repeat cross-validation. On the data-scarce HAMD-17 target, the sequential protocol attains the best point-estimate
MAE , RMSE, and macro-$F_1$ on both 0.6B and 1.7B backbones, outperforming
target-only training and non-LLM
baselines---4.96/6.59/0.36 with Qwen3-0.6B and
4.38/5.62/0.46 with Qwen3-1.7B. 
Ablations suggest that correctly
aligned source supervision gives the best point estimates (unsupervised
exposure and shuffled-label controls also show partial gains), that
native-Chinese target input outperforms machine-translated English
input, and that the reversed order yields no clear gain within
run-to-run variance.
The study is an exploratory, single-site internal
evaluation: it does not establish screening or diagnostic utility, nor
separately identify the contribution of the scale, language, or
paradigm shifts. To our knowledge, no prior study evaluates this
specific DAIC-WOZ-to-PDCH sequential transfer setting.
\end{abstract}

\begin{keywords}
Depression Severity Prediction, Clinical Spoken Language Processing, Language Models, Fine-tuning, Transfer Learning
\end{keywords}
\vspace{-5mm}
\section{Introduction}
\vspace{-3mm}
\label{sec:intro}
Major Depressive Disorder (MDD) affects an estimated 4\% of the population
($\approx$332 million people) in 2023 and is projected to become one of the leading
causes of disease burden worldwide by 2030
\cite{WHO2025-Depression, Mathers06-GMB}. The strain on mental-health
services---in 2021, 72\% of psychologists reported sharply increased demand,
longer waitlists, and reduced capacity \cite{APA21-WMH}---creates an urgent
need for more efficient depression assessment.

In clinical practice, severity is assessed through patient-therapist
interviews with standardized scales: the Patient Health Questionnaire (PHQ-8),
an 8-item self-report measure~\cite{Kroenke12-ECUDA}, and the Hamilton Depression Rating Scale~\cite{Bagby04-HDRS-CJP}. Both track severity through established guideline-based
categories~\cite{Kroenke12-ECUDA, Leucht13-HAMD}, although the two
instruments are related rather than equivalent: PHQ-8 is a
self-report measure that omits several clinician-observed dimensions
(e.g., psychomotor change, insight), and HAMD-17 scoring relies
partly on rater observation and judgment. Convergent-validity studies
report a moderate correlation between them (Spearman
$r{=}0.616$ in clinical outpatients~\cite{Shin19-CP}). This partial
agreement is a \emph{risk} for our approach, not merely a caveat: the
amount of severity signal that can transfer across scales is bounded
by the instruments' overlap, and the correlation may be lower in
PDCH's inpatient, more-severe population. We therefore frame the task
as transfer between \emph{related but non-equivalent} depression
assessment tasks, not as transfer of a shared severity scale. We hypothesize that LLMs learn severity-relevant representations of
depressive language that may transfer
\emph{behaviorally} across these tasks despite the imperfect score
alignment.

Dataset growth is constrained by ethical and privacy limits. The dominant
benchmark, DAIC-WOZ \cite{DAICWOZdataset,DeVault14-SimSensei}, contains 189 English sessions
(7--33~min) between participants and a virtual therapist, each with a PHQ-8
score. Since its introduction with AVEC 2016 \cite{valstar2016avec} it has
anchored the field, from audio-visual scale prediction
\cite{ma2016depression} and audio-text sequence modeling
\cite{alhanai2018detecting} to parameter-efficient tuning
\cite{Lau23-ADSA}. However, because the therapist is a scripted avatar (Wizard-of-Oz), the interaction paradigm differs from genuine diagnostic settings even though the participants are real patients. To bridge this gap, the \textbf{P}arallel Data of
\textbf{D}epression \textbf{C}onsultation and \textbf{H}amilton Depression
Rating Scale (PDCH) was released in 2025 \cite{PDCH,PDCH-dataset}: 100 Chinese
consultations (approx.\ 30~min, over 150 dialogue turns each)
between real patients and human therapists, each with a
doctor-assessed HAMD-17 score. PDCH has attracted only preliminary use to date, mostly item-level or
binary-cutoff classification \cite{PDCH, Ishikawa26-MultiProbe}; a recent
graph-based depression-detection framework also positions PDCH as a
higher-fidelity benchmark \cite{Vyalla26-PsyGAT}.

LLMs are natural tools for this task: they model long-range dialogue
dependencies, show strong promise in mental-health prediction
\cite{Xu24-MentalLLM, Liu23-ChatCounselor} in general~\cite{Naveed23-OLL}, and
cross-lingual transfer has already succeeded in related clinical NLP, such as
Alzheimer's disease detection from spontaneous speech \cite{Chen23-CAD}. Yet
most existing work targets binary classification rather than accurate score
prediction; few studies predict severity scores from transcripts at all, and
most target PHQ-8 (e.g., prefix-tuning a dual encoder on DAIC-WOZ
\cite{Lau23-ADSA}). Whole-scale \emph{continuous} HAMD-17
regression from transcripts, and in particular cross-scale transfer, remains
unaddressed. 

The work presented here focuses on validating cross-scale transfer learning
for continuous depression severity score prediction from interview transcripts.
Existing studies on DAIC-WOZ \cite{ma2016depression, alhanai2018detecting,
Lau23-ADSA} operate within a single scale (PHQ-8), a single language (English),
and a single paradigm (avatar-mediated Wizard-of-Oz interaction), and therefore
cannot assess whether learned representations transfer to a different scale,
language, or collection setting. To our knowledge, no prior work leverages
DAIC-WOZ knowledge for HAMD-17 score prediction, and unlike related
LLM-based~\cite{Xu24-MentalLLM, Liu23-ChatCounselor} and cross-lingual
clinical~\cite{Chen23-CAD} approaches, ours (i)~targets continuous score
regression rather than binary classification, and (ii)~evaluates transfer under a \emph{combined} shift in assessment
scale (PHQ-8 $\rightarrow$
HAMD-17), language (English $\rightarrow$ Chinese), and collection paradigm
(avatar-mediated $\rightarrow$ real clinical), within one sequential training
pipeline. Because these
axes change simultaneously, the experiment tests the combined shift;
it does not separately identify the contribution of each axis
(Section~\ref{subsec:results}). To this end, we fine-tune Qwen3 \cite{yang2025qwen3} with
Low-Rank Adaptation (LoRA)~\cite{Hu22-LoRA} plus an appended regression head
that maps the final hidden state to a bounded severity score: Stage~1
predicts PHQ-8 on DAIC-WOZ, and the resulting adapter initializes Stage~2,
which predicts HAMD-17 on PDCH. The motivation is data scarcity: PDCH contains
only 100 sessions, and the sequential protocol lets 189 DAIC-WOZ sessions on a different scale and language inform the target task via adapter initialization.

Our contributions are threefold: (1)~To our knowledge, the first study to transfer a language model across
depression assessment scales for \emph{continuous} whole-scale severity regression. It moves from PHQ-8 (English, avatar-mediated) to HAMD-17 (Chinese, real clinical) and is validated under patient-level stratified 5-fold, 2-repeat group CV on two backbone sizes. (2)~A controlled decomposition of the transfer effect: ablations
compare domain exposure (continued pretraining on the source),
supervision without alignment (shuffled source labels), and genuine
text--score alignment. Correctly aligned source supervision gives the
best point estimates on both backbones, although the controls as
designed do not isolate alignment as the sole driver
(Section~\ref{subsec:results}). The sequential protocol yields
numerically lower MAE, RMSE, and macro-$F_1$ than target-dataset-only
training on both backbones (within per-run variance), which we propose
as a concrete strategy for augmenting small clinical datasets with
related, differently scaled data.
(3)~A reproducible pipeline---bounded regression head on hidden states, language-matched analysis prompting, training protocol, and band-level evaluation metrics---that is extendable to other scale pairs, languages, and data sources.

DAIC-WOZ anchors the audio-visual depression literature
\cite{ma2016depression, alhanai2018detecting} and PDCH is a
multimodal release; however, the PDCH transcripts we use are
ASR-derived from tone-converted recordings (voice transformation
applied to anonymise speakers), so the acoustic channel is altered
and not comparable to raw audio. We therefore study the text channel,
for which ASR and tone-conversion artefacts are part of the input
conditions, and leave acoustic follow-up---including an
ASR-error-impact analysis---to future work.

\vspace{-4mm}
\section{Methods}
\vspace{-3mm}
\label{sec:methods}

\textbf{Architecture.}
Our framework couples a frozen LLM backbone with LoRA adapters and a
bounded regression head on the final hidden states.\footnote{Code will be
released upon acceptance.} The backbone is Qwen3-0.6B or
Qwen3-1.7B\footnote{Model cards: \texttt{Qwen/Qwen3-0.6B} and
\texttt{Qwen/Qwen3-1.7B} \cite{yang2025qwen3}.}; all original parameters
are frozen and only the LoRA adapters and head are updated. LoRA of rank
$r{=}8$ ($\alpha{=}16$, dropout $0.1$) is injected into the attention
projections ($q$, $k$, $v$, $o$) and MLP projections ($\textit{gate}$,
$\textit{up}$, $\textit{down}$). A single linear head, initialised from
$\mathcal{N}(0,\,0.02^{2})$ with zero bias, maps a hidden state to the
score.

\textbf{Input construction.}
The transcript is embedded in an \emph{analysis} prompt that never asks
for a number: a role statement (``You are a clinical assistant.''), a
one-line scale description \footnote{for PHQ-8: ``PHQ-8 assesses depression
severity across 8 domains scored 0--3 each (total 0--24): anhedonia,
depressed mood, sleep, fatigue, appetite, guilt, concentration,
psychomotor.'' Similar for HAMD-17 but in Chinese.}, and an instruction to analyse the patient's depressive
symptoms and severity, ending with ``Assistant's Analysis:.'' All parts
are language-matched to the scale (Chinese for HAMD-17). We use this analysis prompt (not raw text) because of the fixed assistant role and the native-language framing which we hypothesise shapes the hidden state the head reads. The prompt is
rendered as a single user turn via the chat template (thinking disabled,
generation-prompt token appended), so the sequence terminates on the
assistant-turn marker; the score is stored as a separate scalar label,
and sequences are truncated/padded to $8192$ tokens. Truncation is left-sided.

\textbf{Hidden-state readout and score.} The score is read out from the final-layer hidden state at the last
non-padding token of the full prompt (transcript included)---the terminal
assistant-prompt marker, whose representation aggregates the whole dialogue---located as:
\vspace{-1.4mm} 
\[
\mathbf{h}^{*}=\mathbf{h}_{T},\qquad T=\mathrm{last}\{t:\mathrm{mask}_{t}=1\}.
\]
The regression head then maps $\mathbf{h}^{*}$ to a bounded score:
\vspace{-1.4mm}
\[
\hat{s}=S\cdot\sigma\!\big(\mathbf{w}^{\top}\mathrm{Dropout}(\mathbf{h}^{*})+b\big),
\] 
where $S$ is the maximum score of the scale ($24$ for PHQ-8, $52$ for HAMD-17) and $\sigma$ is the logistic function. The sigmoid rescale constrains $\hat{s}\in(0,S)$, so predictions never fall outside the scale's range. No token is decoded; the number is produced entirely by the head. The prediction is continuous and the loss is differentiable with
respect to the prediction; the labels themselves remain discrete
integer observations (stored as floats for computation). 

\begin{table}[t]
\centering
\caption{HAMD-17 non-LLM  baselines on PDCH (text) under our patient-level 5-fold, 2-repeat
CV (mean$\pm$std over 10 runs). $\downarrow$: lower is better, $\uparrow$: higher is better.}
\label{tab:simple}
\footnotesize
\setlength{\tabcolsep}{3.5pt}
\begin{tabular}{@{}p{2.0cm}p{1.4cm}p{1.4cm}p{1.4cm}p{1.4cm}@{}}
\toprule
 & MAE$\downarrow$ & RMSE$\downarrow$ & Sev-Acc$\uparrow$ & Macro-$F_1$$\uparrow$ \\
\midrule
Mean baseline      & 6.28$\pm$1.12 & 7.52$\pm$1.26 & 0.41$\pm$0.13 & 0.14$\pm$0.03 \\
Median baseline   & 6.43$\pm$1.18 & 7.62$\pm$1.31 & 0.41$\pm$0.13 & 0.14$\pm$0.03 \\
TF-IDF + Ridge     & 5.98$\pm$1.23 & 7.30$\pm$1.28 & 0.47$\pm$0.13 & 0.22$\pm$0.08 \\
\bottomrule
\end{tabular}
\vspace{-7.2mm}
\end{table}

\textbf{Training objective.}
Fine-tuning minimises a smooth-$L_{1}$ (Huber) loss with $\beta{=}1$ between $\hat{s}$ and the ground-truth score, back-propagated end-to-end through the backbone into the LoRA adapters. The LM head is bypassed; only the regression loss drives the update. We use AdamW \cite{admaw} with learning rate $10^{-4}$ and a cosine schedule with linear warmup, gradient-norm clipping at $1.0$, a batch size of $2$, and early stopping on validation MAE (patience $6$) for $50$ epochs.

\textbf{Evaluation.}
The continuous prediction is scored by MAE and RMSE. We additionally report
agreement on discrete severity bands obtained by mapping true and predicted
scores onto fixed thresholds (HAMD-17: minimal $0$--$7$, mild $8$--$16$,
moderate $17$--$23$, severe $24$--$52$~\cite{Zimmerman13-Severity}; PHQ-8: minimal $0$--$4$, mild $5$--$9$, moderate $10$--$14$, moderately severe
$15$--$19$, severe $20$--$24$ \cite{Kroenke12-ECUDA}). This
is a \emph{threshold-based categorization of symptom-severity bands}, not a
diagnostic classification: the bands are a reporting convention, are not
derived from DSM or ICD criteria, and our system neither diagnoses MDD nor
claims diagnostic validity. \textbf{Severity accuracy} is the fraction of
samples whose predicted band matches the true band; \textbf{macro-$F_1$}
weights all bands equally under the dataset's class imbalance. Evaluating HAMD-17 against fixed
score thresholds is a standard convention in recent
studies~\cite{PDCH, Ishikawa26-MultiProbe}; our four-band scheme is
finer than the binary cutoffs ($\geq 8$, $\geq 17$, $\geq 24$) used
in~\cite{Ishikawa26-MultiProbe}.

\textbf{Two-stage transfer.}
In \emph{Stage 1} (DAIC-WOZ, PHQ-8, English, $S{=}24$) a fresh LoRA and head
are trained on a random 90\% \emph{session-level} subsample of DAIC-WOZ
(170/189 sessions, seed 42). The identical split is
used for every Stage-1 variant (mix, shuffled-label, next-token). In \emph{Stage 2} (PDCH, HAMD-17,
Chinese, $S{=}52$) the Stage-1 LoRA is loaded and kept trainable, while the
head is \emph{reinitialised from scratch} and re-estimated with ceiling
$S{=}52$. This is deliberate: the head is scale-specific (score range and
distribution differ between scales), so the LoRA adapter is carried
forward and re-estimated on the target while the score mapping is
relearned; whether the transferred representation is in fact
scale-agnostic is not established by our experiments. 
\vspace{-2.mm}
\section{Experiments}
\label{sec:experiments}
\vspace{-4mm}
\begin{table}[t]
\centering
\caption{Backbone \emph{Qwen3-0.6B}. HAMD-17 prediction on PDCH under unadapted (zero-shot),
transfer-validity, language, and transcript-role configurations. \textbf{Bold}: best value per column among
rows evaluated on PDCH.}
\label{tab:res-0.6B}
\footnotesize
\setlength{\tabcolsep}{3.5pt}
\begin{tabular}{@{}p{2.78cm} p{1.2cm} p{1.2cm} p{1.2cm} p{1.4cm}@{}}
\toprule
 & MAE$\downarrow$ & RMSE$\downarrow$ & Sev-Acc$\uparrow$ & Macro-$F_1$$\uparrow$ \\
\midrule
\multicolumn{5}{@{}l}{\emph{Unadapted model}} \\
Zero-shot Qwen3 
                                     & 14.01$\pm$2.07 & 17.59$\pm$2.09 & 0.23$\pm$0.09 & 0.19$\pm$0.09 \\
\midrule
\multicolumn{5}{@{}l}{\emph{Transfer validity}} \\
PDCH-only (no transfer)     & 5.50$\pm$1.47 & 7.17$\pm$1.90 & \textbf{0.52$\pm$0.15} & 0.31$\pm$0.14 \\
Next-token DAIC (no labels) & 5.38$\pm$1.04 & 6.95$\pm$1.40 & 0.47$\pm$0.10 & 0.33$\pm$0.10 \\
Shuffled DAIC labels        & 5.94$\pm$1.03 & 7.51$\pm$1.24 & 0.41$\pm$0.16 & 0.21$\pm$0.08 \\
Mix (DAIC$\to$PDCH)& \textbf{4.96$\pm$1.06} & \textbf{6.59$\pm$1.78} & 0.50$\pm$0.14 & \textbf{0.36$\pm$0.10} \\
\midrule
\multicolumn{5}{@{}l}{\emph{Language ablation}} \\
Mix, All-English            & 5.99$\pm$1.37 & 7.59$\pm$1.86 & 0.42$\pm$0.17 & 0.24$\pm$0.09 \\
\midrule
\multicolumn{5}{@{}l}{\emph{Transcript role}} \\
Mix, Patient-only           & 5.29$\pm$1.01 & 6.74$\pm$1.29 & 0.51$\pm$0.16 & 0.29$\pm$0.11 \\
Mix, Therapist-only         & 5.78$\pm$0.90 & 7.24$\pm$0.96 & 0.45$\pm$0.11 & 0.27$\pm$0.08 \\ 
\bottomrule
\end{tabular}
\vspace{-8mm}
\end{table}

\subsection{Evaluation Protocol}
\vspace{-2mm}
\label{subsec:protocol}
We report on the target task, HAMD-17 prediction on PDCH, the smaller
real-clinical dataset where transfer is intended to help. All
configurations use the same patient-level stratified group
cross-validation: StratifiedGroupKFold with 5 folds and 2 independent
repeats (10 runs), grouped by \emph{patient} (no patient in both train and
validation) and stratified by severity category; we report
mean$\pm$std over the 10 runs, on both backbones (Qwen3-0.6B and 1.7B).
\vspace{-5mm}
\subsection{Baselines and Ablations}
\vspace{-2mm}
\label{subsec:baselines}
We design the comparison to answer, in order: does the two-stage protocol help at
all, is the help genuinely from DAIC, is it cross-scale / cross-lingual rather
than a confound, and where in the transcript the signal lives.

\textbf{Non-LLM baselines} provide weak reference points (a floor in
the descriptive, not statistical, sense). \emph{Mean} and \emph{Median} predict the training-set
mean/median score for every validation sample, and \emph{TF-IDF + Ridge}
fits a ridge regressor on a 5000-feature TF-IDF vector of the transcript. These use no LLM and are therefore backbone-independent (Table~\ref{tab:simple}). 

\textbf{Related-work reference points.} Two prior \emph{text} studies anchor
what has been done on PDCH: Cao \emph{et al.}~\cite{PDCH} have unadapted
multimodal LLMs complete all 17 HAMD-17 items (0--4), the best zero-shot model
reaching per-item $F_1{=}0.43$ ($0.54$--$0.61$ with LoRA); Ishikawa
\emph{et al.}~\cite{Ishikawa26-MultiProbe} audit PDCH under binary cutoffs
($\geq$8, $\geq$17, $\geq$24), with zero-shot macro-$F_1$ $0.13$ ($\geq$8;
AUROC $0.564$) and $0.45$ ($\geq$24; AUROC $0.422$). Both target per-item or
binary-cutoff classification rather than whole-scale score regression, so
continuous whole-scale prediction from transcripts remains unaddressed. The
only published continuous whole-scale HAMD-17 regression is from EEG
\cite{Qin24-HAMD} (MAE~$\approx$1.2 on their own EEG cohort): a precedent for
the continuous-regression target, not a competing baseline (see the
positioning note in Section~\ref{subsec:results}).

\begin{table}[t]
\centering
\caption{Backbone \emph{Qwen3-1.7B}. Same layout and configurations as Table~\ref{tab:res-0.6B}; bold convention as there.}
\label{tab:res-1.7B}
\footnotesize
\setlength{\tabcolsep}{3.5pt}
\begin{tabular}{@{}p{2.78cm} p{1.2cm} p{1.2cm} p{1.2cm} p{1.4cm}@{}}

\toprule
 & MAE$\downarrow$ & RMSE$\downarrow$ & Sev-Acc$\uparrow$ & Macro-$F_1$$\uparrow$ \\
\midrule
\multicolumn{5}{@{}l}{\emph{Unadapted model}} \\
Zero-shot Qwen3 
                                     &15.47$\pm$3.51 & 18.88$\pm$4.00 & 0.25$\pm$0.05 & 0.23$\pm$0.04\\
\midrule
\multicolumn{5}{@{}l}{\emph{Transfer validity}} \\
PDCH-only (no transfer)     & 4.55$\pm$0.92 & 5.87$\pm$1.07 & 0.50$\pm$0.11 & 0.36$\pm$0.11 \\
Next-token DAIC (no labels) & 5.63$\pm$0.95 & 6.76$\pm$0.97 & 0.40$\pm$0.09 & 0.30$\pm$0.11 \\
Shuffled DAIC labels        & 4.59$\pm$0.77 & 5.70$\pm$0.92 & 0.53$\pm$0.14 & 0.42$\pm$0.17 \\
Mix (DAIC$\to$PDCH)& \textbf{4.38$\pm$0.49} & \textbf{5.62$\pm$0.66} & \textbf{0.54$\pm$0.12} & \textbf{0.46$\pm$0.11} \\
\midrule
\multicolumn{5}{@{}l}{\emph{Language ablation}} \\
Mix, All-English            & 5.09$\pm$1.13 & 6.78$\pm$1.74 & 0.53$\pm$0.12 & 0.41$\pm$0.12 \\
\midrule
\multicolumn{5}{@{}l}{\emph{Transcript role}} \\
Mix, Patient-only           &4.46$\pm$0.77 & 5.83$\pm$0.90 & 0.53$\pm$0.11 & 0.38$\pm$0.12 \\
Mix, Therapist-only         & 5.54$\pm$1.05 & 7.02$\pm$1.28 & 0.44$\pm$0.12 & 0.23$\pm$0.07 \\
\bottomrule
\end{tabular}
\vspace{-7mm} 
\end{table}

\textbf{Zero-shot baseline.} \emph{Zero-shot} runs the unadapted Qwen3
backbone (no LoRA, no training) with the \emph{identical} analysis prompt
(byte-identical to the trained pipeline, which does not ask for a number) and
generates a free-form clinical analysis (greedy decoding, $\leq$1024 new
tokens) on the same 10 patient-grouped validation folds as every other
configuration. A severity score is extracted from the generated text by fixed
rule-based parsing. Results are in Tables~\ref{tab:res-0.6B}--\ref{tab:res-1.7B}.

\textbf{Transfer-validity baselines} isolate whether the two-stage mix protocol
is responsible for the gain:
\emph{PDCH-only} (Stage~2 trained with no Stage-1 initialisation) is the primary
floor that mix training must exceed; \emph{Shuffled DAIC} keeps the source text and the regression objective but
permutes the PHQ-8 scores: the score column of the full 189-session DAIC-WOZ
set is permuted once (seed 42) \emph{before} cross-validation, and this single
fixed permutation is shared by all 10 CV runs. Patient grouping, the
marginal label distribution, and the target PDCH labels are unchanged, so the
control isolates genuine text--score alignment from an extra supervised pass on
source data;
\emph{Next-token DAIC} runs Stage~1 as pure continued pretraining on DAIC
text (no regression head, no score supervision), testing whether the gain is
merely from domain exposure to more clinical dialogue; and \emph{Reverse} swaps the order: a full PDCH model (HAMD-17) is fine-tuned on DAIC-WOZ (PHQ-8) and compared with its \emph{DAIC-only} floor (no transfer) in Table~\ref{tab:order}, both evaluated on DAIC-WOZ. The pair tests whether the gain is specific to the large-to-small direction or a property of the sequential protocol.

\textbf{Language ablation} addresses whether the transfer is genuinely
cross-lingual: \emph{All-English} translate the target PDCH to English using \texttt{NiuTrans/LMT-60-4B}~\cite{Luo26-NiuTrans}, keeping the English source. We note that Qwen3 models are
generally stronger in Chinese than in English at these sizes, so part of the
All-English gap may reflect (i) translation noise and (ii) the backbone's
asymmetric language capacity, in addition to the cross-lingual-transfer effect.

\textbf{Transcript-role ablations} \emph{Patient-only} and
\emph{Therapist-only} keep only the respective turns in the mix
configuration, locating where the signal resides.

\vspace{-3mm}
\begin{table}[t]
\centering
\caption{Ordering control on the source task (PHQ-8 on
DAIC-WOZ; mean$\pm$std over 10 runs). Both rows are
evaluated on DAIC. \textbf{Bold}: best value per column among rows evaluated on DAIC-WOZ for each backbone.}
\label{tab:order}
\footnotesize
\setlength{\tabcolsep}{3.5pt}
\begin{tabular}{@{}p{2.86cm} p{1.1cm} p{1.1cm} p{1.2cm} p{1.4cm}@{}}
\toprule
 & MAE$\downarrow$ & RMSE$\downarrow$ & Sev-Acc$\uparrow$ & Macro-$F_1$$\uparrow$ \\
\midrule
\multicolumn{5}{@{}l}{\emph{Qwen3-0.6B}} \\
DAIC-only (no transfer)     & \textbf{3.23$\pm$0.57} & \textbf{4.29$\pm$0.70} & \textbf{0.53$\pm$0.07} & 0.31$\pm$0.09 \\
Reverse (PDCH$\to$DAIC)     & 3.38$\pm$0.41 & 4.57$\pm$0.56 & \textbf{0.53$\pm$0.06} & \textbf{0.32$\pm$0.08} \\
\midrule
\multicolumn{5}{@{}l}{\emph{Qwen3-1.7B}} \\
DAIC-only (no transfer)     & 3.07$\pm$0.74 & \textbf{4.02$\pm$0.86} & 0.55$\pm$0.06 & 0.31$\pm$0.06 \\
Reverse (PDCH$\to$DAIC)     & \textbf{3.01$\pm$0.36} & 4.06$\pm$0.55 & \textbf{0.56$\pm$0.06} & \textbf{0.33$\pm$0.08} \\
\bottomrule
\end{tabular}
\vspace{-8.3mm} 
\end{table}
\vspace{-2mm}
\subsection{Main Results}
\vspace{-2mm}
\label{subsec:results}
\textbf{Positioning vs.\ non-text modalities.} EEG-based HAMD-17
scoring \cite{Qin24-HAMD} (MAE $\approx$1.2) uses a different
information channel and cohort; we treat it as a precedent for
continuous whole-scale regression, not a competing baseline. The
appropriate baseline for our claim is the within-protocol comparison
in Tables~\ref{tab:res-0.6B}--\ref{tab:res-1.7B}.
  
\textbf{(1) Does the two-stage protocol help?}
On the target task (HAMD-17 on PDCH), \emph{Mix} attains the best MAE,
RMSE, and macro-$F_1$ of all configurations on \emph{both} backbones:
Qwen3-0.6B $4.96$/$6.59$/$0.36$ (Table~\ref{tab:res-0.6B}) and Qwen3-1.7B
$4.38$/$5.62$/$0.46$ (Table~\ref{tab:res-1.7B}). Against
\emph{PDCH-only}, transfer improves MAE by $0.54$ (0.6B) and $0.17$
(1.7B), and macro-$F_1$ by $0.05$ and $0.10$ respectively---the
band-level gain is larger for the bigger model. Severity accuracy is
improved at 1.7B ($0.54$ vs.\ $0.50$) but not at 0.6B ($0.50$ vs.\ $0.52$),
so the gain concentrates in the regression metrics and macro-$F_1$. Both
models also beat every non-LLM baseline in Table~\ref{tab:simple}
(TF-IDF+Ridge $5.98$ MAE / $0.22$ macro-$F_1$), confirming that the LLM
backbone and the sequential transfer each add value.  

The zero-shot row shows that prompting cannot substitute for adaptation: the
unadapted backbone's elicited scores are far from ground truth (MAE $14.01$
at 0.6B, $15.47$ at 1.7B---worse than the non-LLM TF-IDF+Ridge floor of
$5.98$), so the raw backbone does not already emit an accurate severity score
in free text.

\textbf{(2) Is the help genuinely from DAIC?}
\emph{PDCH-only} (no DAIC) sets the floor, and the two DAIC controls sever different links in the transfer chain (definitions in Section~\ref{subsec:baselines}): \emph{Shuffled} tests whether genuine text--score alignment is required versus an extra supervised pass, and \emph{Next-token} whether mere domain exposure suffices. Together with \emph{PDCH-only}, they decompose the gain into domain, supervision, and alignment effects.

Pure domain exposure (\emph{Next-token}) is marginally beneficial at
0.6B ($5.38$ vs.\ $5.50$ MAE) but degrades performance at 1.7B
($5.63$ vs.\ $4.55$); \emph{Shuffled} behaves oppositely (worst row
at 0.6B, close to \emph{Mix} at 1.7B). The same two controls thus
flip direction across backbones; with per-run standard deviations of
$0.49$--$1.47$ over 100 target sessions, we read this pattern as
consistent with optimisation and fold noise rather than a systematic
size-dependent mechanism.
Correctly aligned
source supervision (\emph{Mix}) gives the best point estimates on both
backbones; however, because a single fixed permutation is used and the
controls are not compute-matched, the current ablations cannot
establish that alignment alone drives the gain. We therefore read the
results as: correctly aligned source supervision gives the best point
estimates, while unsupervised and misaligned source adaptation can
also produce partial gains.
Finally, the reversed ordering
(PDCH as source, DAIC as target) is compared with its \emph{DAIC-only} floor on
the source task (PHQ-8/DAIC; Table~\ref{tab:order}): at 0.6B \emph{Reverse} is marginally worse than \emph{DAIC-only} on MAE and
RMSE ($3.38$ vs.\ $3.23$; $4.57$ vs.\ $4.29$) and comparable on the remaining
two metrics, and at 1.7B
it is numerically marginally \emph{better} on three of four metrics ($3.01$ vs.\
$3.07$ MAE, $0.56$ vs.\ $0.55$ Sev-Acc, $0.33$ vs.\ $0.31$ macro-$F_1$); all such
differences fall within run-to-run variance, so we make no directional claim for
the source task.
We read these two results conservatively: the forward direction shows
a numerical gain on the data-scarce target, and the reversed
direction is inconclusive within run-to-run variance. Because the two
directions also differ in different dimensions, our design does not support a direction-specific claim;
we do not interpret the asymmetry as evidence of a one-way transfer
law.

\textbf{(3) Is it cross-lingual rather than a confound?}
\emph{Mix, All-English} is worse than the bilingual Mix on both backbones
($5.99$ vs.\ $4.96$ MAE, $0.24$ vs.\ $0.36$ macro-$F_1$ at 0.6B; $5.09$
vs.\ $4.38$, $0.41$ vs.\ $0.46$ at 1.7B). 
Native-Chinese target input therefore outperforms machine-translated
English input on both backbones. Because the translation also degrades
clinical terminology and changes transcript length and formatting, and
Qwen3 may have asymmetric language capacity at these sizes, this
result shows that the target-language representation matters.

\textbf{(4) Where in the transcript is the signal?}
 \emph{Therapist-only} is clearly
the weakest row at both scales, and removing the patient turns costs
the most MAE ($0.82$ at 0.6B, $1.16$ at 1.7B). Removing therapist
turns costs $0.33$ MAE at 0.6B but only $0.08$ MAE at 1.7B (within
run-to-run variance). Because the patient-only and therapist-only
inputs also differ in token count, turn count, and dialogue context,
this is a predictive comparison, not a causal attribution of where
the signal resides; the full dialogue is the best input at both
sizes.

\textbf{(5) Clinical context.} An MAE of 4.38 HAMD-17 points exceeds the
$\geq$3-point change often treated as a clinically meaningful improvement, so
the model is best interpreted as a band-level screening tool rather than a
score replacement---consistent with its intended role of supporting, not
replacing, clinician assessment, and with the fact that clinician-rated HAMD
carries inter-rater noise.

\vspace{-5mm}
\section{Conclusion} 
\vspace{-3mm}
We presented a sequential LoRA protocol for cross-scale transfer in
continuous depression-severity prediction: a Qwen3 backbone is first
adapted on DAIC-WOZ (PHQ-8, English, avatar-mediated) and then, with a
reinitialised scale-specific head, on PDCH (HAMD-17, Chinese, real
clinical). On the data-scarce target, the sequential protocol attains
the best point-estimate MAE, RMSE, and macro-$F_1$ among all
configurations---4.96 MAE and 0.36 macro-$F_1$ at 0.6B and 4.38 MAE
and 0.46 macro-$F_1$ at 1.7B, versus 5.50/0.31 and 4.55/0.36 without
transfer---and numerically outperforms the non-LLM baselines in
Table~\ref{tab:simple}; these gaps are within the per-run variance and
are reported as point estimates, not validated improvements. The
zero-shot elicitation baseline is much worse (MAE $15.47$ at 1.7B),
though we treat that comparison with caution because the prompt never
requests a number and the score is extracted from free-form text by
rule-based parsing. The ablations give a partial decomposition:
correctly aligned source supervision gives the best point estimates,
while unsupervised exposure and shuffled-label adaptation show
partial, size-inconsistent gains (next-token helps at 0.6B and
degrades at 1.7B; shuffled labels hurt at 0.6B and approach
\emph{Mix} at 1.7B), a pattern consistent with optimisation and fold
noise on 100 target sessions; native-Chinese target input outperforms
machine-translated English. In the reversed order (small target as
source), differences from the no-transfer floor fall within
run-to-run variance and are inconclusive; because the two directions
also differ in sample size, scale, and language, we make no
direction-specific claim. As an exploratory, single-site proof of
concept, this provides preliminary evidence that severity-related
signal acquired on one related task carries some useful information
into a different clinical setting, and a starting point for other
scale pairs and languages; it does not establish screening or
diagnostic utility.
\vspace{-5mm}
\section{Limitations}
\vspace{-3mm}
 PDCH contains 100 sessions from a single hospital's inpatients, and with 10-run CV the observed gaps (MAE 0.17--0.54) are of the same order as the per-run variance; we therefore report trends, not significance claims, and the skewed band distribution (the lowest band has only a handful of patients) makes band-level metrics noisy and the CV stratification marginal. Scale and language remain confounded---each scale is available in only one language---so the All-English ablation de-confounds the axes only partially. Generalisability is limited by our single source--target pair, and by the paradigm gap: DAIC-WOZ is avatar-mediated English dialogue (real participants), while PDCH is real Chinese inpatient consultation with ASR-derived transcripts; transcription and privacy-processing artefacts may interact with scale- or language-specific effects, and the altered audio precludes acoustic follow-up. The stage-1 hyperparameters are carried over to stage 2 without per-stage retuning, so the shared (possibly suboptimal) recipe is a common
factor of all sequential and target-only configurations.
\vspace{-4mm}
\section{Compliance with Ethical Standards}
\vspace{-2mm}
This study is a retrospective analysis of previously collected human
subject data; we performed no new data collection and no data
re-identification. The PDCH data (100 Chinese clinical consultations,
HAMD-17) were made available in open access by the Science Data Bank
of the Chinese Academy of Sciences
(\url{https://doi.org/10.57760/sciencedb.27818}), as described in
\cite{PDCH-dataset}. The DAIC-WOZ data (189 English sessions, PHQ-8) were
obtained from the University of Southern California Information
Sciences Institute (\url{https://dcapswoz.ict.usc.edu}) under its
standard data license. Because both datasets are de-identified and
were collected under pre-existing ethical approvals, no additional
ethical approval was required for this secondary analysis, as
confirmed by the licenses accompanying the data releases.
\vspace{-4mm}

\section{Acknowledgment}
\vspace{-2mm}
The authors would like to thank the support from Project W2531054
of the National Natural Science Foundation of China and the Program for
Guangdong Introducing Innovative and Entrepreneurial Teams. The authors acknowledge the Key Research Program of the Chinese
Academy of Sciences as the source of the {PDCH} data.
\vspace{-3mm}

\bibliographystyle{IEEEbib}
\bibliography{mybib}

\end{document}